%% file: gecco-en4sr_00_main.tex
\documentclass[sigconf]{acmart}
\AtBeginDocument{%
  }

\copyrightyear{2025} 
\acmYear{2025} 
\setcopyright{cc}
\setcctype{by}
\acmConference[GECCO '25]{Genetic and Evolutionary Computation Conference}{July 14--18, 2025}{Malaga, Spain}
\acmBooktitle{Genetic and Evolutionary Computation Conference (GECCO '25), July 14--18, 2025, Malaga, Spain}\acmDOI{10.1145/3712256.3726434}
\acmISBN{979-8-4007-1465-8/2025/07}

\usepackage[vlined,algoruled,commentsnumbered]{algorithm2e}
\usepackage{subfigure}
\usepackage{arydshln}
\usepackage{caption}
\usepackage{subcaption}
\usepackage{booktabs}
\usepackage{url}            

\newcommand{\ennsr} {\texttt{EN4SR}}
\newcommand{\ennsrBase} {\texttt{EN4SR-base}}
\newcommand{\eql} {$\texttt{EQL}^{\div}$}
\newcommand{\nnsr} {\texttt{N4SR}}
\newcommand{\nnsrACYE} {\texttt{N4SR-ACYE}}
\newcommand{\sngp} {\texttt{mSNGP-LS}}

\newcommand{\mastera} {\texttt{master-A}}
\newcommand{\masterb} {\texttt{master-B}}
\newcommand{\mTopLow} {master topology}

\newcommand{\sTop} {subtopology}
\newcommand{\sTops} {subtopologies}
\newcommand{\sTopCap} {Subtopology}
\newcommand{\basis} {elementary}
\newcommand{\basemodel} {\emph{reference model}}
\newcommand{\resistors} {\texttt{resistors}}
\newcommand{\magman} {\texttt{magman}}
\newcommand{\magic} {\texttt{magic}}

\DeclareMathOperator{\cube}{cube}
\DeclareMathOperator{\ident}{ident}

\let\oldnl\nl
\newcommand{\nonl}{\renewcommand{\nl}{\let\nl\oldnl}}
\newlength\lenKwIn
\newcommand\myKwIn[1]{%
  \settowidth\lenKwIn{\KwIn{}}%
  \setlength\hangindent{\lenKwIn}%
  \nonl\hspace*{\lenKwIn}#1\\}

\newlength\lenKwOut

\SetKwData{sinT}{$T^{s}$}
\SetKwData{conT}{$T^{c}$}
\SetKwData{vecW}{$\mathbf{W}$}
\SetKwData{vecWl}{$W_l$}    
\SetKwData{vecWs}{$W_s$}    
\SetKwData{vecd}{$\mathbf{d}$}
\SetKwData{vecu}{$\mathbf{u}$}
\SetKwData{vecU}{$U$}       
\SetKwData{vecUl}{$U_l$}    
\SetKwData{vecUc}{$U_c$}    
\SetKwData{vecx}{$\mathbf{x}$}
\SetKwData{vecy}{$\mathbf{y}$}
\SetKwData{thetaa}{$\theta^{a}$}    
\SetKwData{D}{$D$}  
\SetKwData{Dt}{$D_t$}  
\SetKwData{Dv}{$D_v$}  
\SetKwData{Dc}{$D_c$}  
\SetKwData{Di}{$D_i$}  
\SetKwData{De}{$D_e$}  
\SetKwData{domWhole}{$\mathbb{D}_o$}  
\SetKwData{domTrain}{$\mathbb{D}_t$}  
\SetKwData{domExt}{$\mathbb{D}_e$}    
\SetKwData{domInt}{$\mathbb{D}_i$}    
\SetKwData{Debar}{$\bar{D}_e$}  
\SetKwData{LI}{$\mathcal{L}_{1}$}  
\SetKwData{LII}{$\mathcal{L}_{2}$}  
\SetKwData{LIII}{$\mathcal{L}_{3}$}  
\SetKwData{Lsu}{$\mathcal{L}^{s}$}  
\SetKwData{Ltr}{$\mathcal{L}^{t}$}  
\SetKwData{Lreg}{$\mathcal{L}^{r}$}  
\SetKwData{Lcve}{$\mathcal{L}^{c}$}  
\SetKwData{craw}{$\rho^c_{j,k}$}
\SetKwData{nI}{$N_{init}$}   
\SetKwData{nIII}{$N_{final}$} 
\SetKwData{nexp}{$N_{e}$}  
\SetKwData{nfoc}{$N_{f}$}  
\SetKwData{E}{$E$}  
\SetKwData{window}{$N_{w}$}  
\SetKwData{sngtrain}{$r_{s/t}$}  
\SetKwData{cvetrain}{$r_{c/t}$}  
\SetKwData{regtrain}{$r_{r/t}$}  

\SetKwFunction{isactive}{isActive}
\SetKwFunction{getconfig}{getConfig}
\SetKwFunction{randomconfig}{randomConfig}
\SetKwFunction{notempty}{notEmpty}
\SetKwFunction{init}{init}
\SetKwFunction{update}{update}
\SetKwFunction{merge}{merge}
\SetKwFunction{perturb}{perturb}
\SetKwFunction{getnondom}{getNonDominated}
\SetKwFunction{sortnondom}{nonDominatedSorting}
\SetKwFunction{runsgd}{runSGD}
\SetKwFunction{runnsga}{generateSubtopologies}
\SetKwFunction{merge}{merge}
\SetKwFunction{selectbest}{selectBest}

\begin{document}

\title{Neuro-Evolutionary Approach to Physics-Aware Symbolic Regression}


\author{Ji\v{r}\'i Kubal\'ik}
\orcid{orcid_1}
\affiliation{
  Czech Institute of Informatics, Robotics, and Cybernetics CTU in Prague
  \city{Prague}
  \country{Czech Republic}
}
\email{jiri.kubalik@cvut.cz}

\author{Robert Babu{\v s}ka}
\orcid{orcid_2}
\affiliation{
  Dept.\ of Cognitive Robotics, TU Delft, The~Netherlands
  \city{CIIRC, CTU in Prague, Czech Republic}
  \country{}
}
\email{r.babuska@tudelft.nl}

\renewcommand{\shortauthors}{ }

\begin{abstract}
\input{gecco-en4sr_abstract}
\end{abstract}




\maketitle

\input{gecco-en4sr_introduction}

\input{gecco-en4sr_related_work}

\input{gecco-en4sr_problem}

\input{gecco-en4sr_method}

\input{gecco-en4sr_experiments}

\input{gecco-en4sr_results}

\input{gecco-en4sr_conclusions}


\begin{acks}
  This work has received funding from the European Union’s Horizon Europe research and innovation programme under grant agreement No. 101070254 CORESENSE. Views and opinions expressed are, however, those of the authors only and do not necessarily reflect those of the European Union or the Horizon Europe programme. Neither the European Union nor the granting authority can be held responsible for them.
  This work was also partly supported by the European Union under the project ROBOPROX (reg. no. CZ.02.01.01/00/22\_008/0004590).
\end{acks}





\bibliographystyle{ACM-Reference-Format}
\bibliography{gecco-en4sr}





\end{document}

%% file: gecco-en4sr_abstract.tex
Symbolic regression is a technique that can automatically derive analytic models from data. Traditionally, symbolic regression has been implemented primarily through genetic programming that evolves populations of candidate solutions sampled by genetic operators, crossover and mutation. More recently, neural networks have been employed to learn the entire analytical model, i.e., its structure and coefficients, using regularized gradient-based optimization. Although this approach tunes the model's coefficients better, it is prone to premature convergence to suboptimal model structures.
Here, we propose a neuro-evolutionary symbolic regression method that combines the strengths of evolutionary-based search for optimal neural network (NN) topologies with gradient-based tuning of the network's parameters.
Due to the inherent high computational demand of evolutionary algorithms, it is not feasible to learn the parameters of every candidate NN topology to the full convergence. Thus, our method employs a memory-based strategy and population perturbations to enhance exploitation and reduce the risk of being trapped in suboptimal NNs. 
In this way, each NN topology can be trained using only a short sequence of backpropagation iterations.
The proposed method was experimentally evaluated on three real-world test problems and has been shown to outperform other NN-based approaches regarding the quality of the models obtained.

%% file: gecco-en4sr_introduction.tex
\section{Introduction}
\label{sec:intro}

Symbolic regression (SR) is a data-driven technique that creates models in the form of mathematical expressions. It has achieved remarkable results in various nonlinear modeling problems \cite{Schmidt2009, Richmond2017, Udrescu2020, Derner2020}. Traditionally, SR has been implemented using genetic programming (GP) \cite{Schmidt2009,Koza1992,Staelens2013,Arnaldo2015,Bladek2019,Burlacu2020OperonCA}. An evolutionary algorithm refines a population of candidate solutions over multiple generations, guided by a selection mechanism that favors high-quality solutions over weaker ones. New candidate models are generated through genetic operations such as crossover and mutation. More recent GP-based methods also incorporate the loss function gradient to fine-tune the model's internal coefficients \cite{topchy2001,zegklitz2019,Kommenda2020}.

Symbolic regression offers several benefits compared to other data-driven modeling techniques. Unlike deep neural networks, which typically require large amounts of data to perform well, SR can generate accurate models even from a limited number of training samples \cite{Wilstrup2021SymbolicRO,Derner21}. Furthermore, SR is well suited to integrate prior knowledge about the system being modeled \cite{Bladek2019,Kubalik2020,Kubalik2021eswa}. This is valuable when the dataset does not uniformly cover the input space or when there are no data samples in certain regions of the input space. Even with a sufficiently large and diverse dataset, methods that focus solely on training error minimization can produce inaccurate models. For instance, in the case of dynamic system models, errors typically occur in steady-state behavior or incorrect models of local dynamics.

Although GP-based symbolic regression methods are widely used, they have several drawbacks. One major issue is the tendency of the model to grow in size and complexity without a corresponding performance improvement, a phenomenon referred to as code bloat \cite{Trujillo2016}. In addition, these methods become inefficient for large numbers of variables and samples in the training dataset. This is because they continuously evolve and evaluate an entire population of formulas. Finally, fine-tuning the model coefficients solely through genetic operators proved to be quite challenging for the pure GP methods.

To address the challenges outlined above, various methods have recently been developed that leverage neural networks (NNs) to learn analytic formulas through gradient-based optimization techniques \cite{martius2016, sahoo2018, werner2021, Costa2021, Kim2021, Zhou2022}. These methods share a common principle: analytic models are represented by a heterogeneous NN, whose units perform mathematical operations and elementary functions such as \{+, -, *, /, sin, exp, etc.\}. Network weights are optimized using standard gradient-based algorithms to minimize the training error while significantly reducing the number of active units, i.e., the units that contribute to the output, through regularization. The resulting NN then represents a parsimonious analytic formula. Individual approaches differ primarily in how they guide the learning process toward the final model.
Another NN-based symbolic regression method for constructing physically plausible models using minimal training data and prior system knowledge was introduced in \cite{kubalik_n4sr_2023}. It employs an adaptive weighting scheme for balancing multiple loss terms and an epoch-wise learning strategy to minimize the risk of getting stuck in suboptimal local optima. Unlike the aforementioned NN-based approaches, this one incorporates prior knowledge about the model sought, making it more robust for practical deployment in real-world applications. 

These NN-based approaches also have notable drawbacks: they are inefficient in reducing the initial redundant NN architecture to a final sparse form, and gradient-based learning is inherently prone to premature convergence to suboptimal topologies, making it difficult to transition to a better one.

This paper introduces a novel neuro-evolutionary SR algorithm, \ennsr, which integrates the advantages of the two main SR approaches mentioned above. This method combines an evolutionary search for optimal neural network topologies with gradient descent-based optimization of the network weights. In addition, it employs a technique we term {\em weight memory} and regular population perturbation to improve exploration and minimize the likelihood of getting stuck in suboptimal neural network topologies. The candidate NN models generated through this process adopt the architecture and the loss functions used in gradient-based learning inspired by those introduced for the \nnsr\ method \cite{kubalik_n4sr_2023}.

The main challenge is to efficiently sample and refine candidate NN topologies despite the increased computational demands that naturally arise from the nature of evolutionary algorithms. The contributions of the paper are:
\begin{itemize}
    \item Design of a neuro-evolutionary SR method that combines advantages of the gradient-free evolutionary approach to generate candidate NN topologies and the gradient-based tuning of the NN weights.
    \item Design of a memory-based strategy to facilitate effective use of crossover and mutation operators regarding the weights of NN structures modified by these operators.
    \item The \ennsr\ method proved to be as good as or better than the pure NN-based approach in terms of model accuracy and computational demands while outperforming the other method in the frequency of produced high-quality solutions.
    \item The \ennsr\ method has been shown to significantly outperform the GP-based approach in terms of the quality of generated models.
\end{itemize}

The \ennsr\ algorithm was evaluated on four test problems and compared to other existing methods, among which is the NN-based \nnsr\ serving as the baseline for comparisons with our neuro-evolutionary approach. In addition, an ablation study demonstrated the effect of some of the essential components of the algorithm.

The paper is organized as follows. Section~\ref{sec:related} surveys the related work. Then, Section~\ref{sec:problem} outlines the particular SR problem considered in this work. The proposed method is described in Section~\ref{sec:method} and Section~\ref{sec:experiments} presents the experiments and the results obtained. Finally, Section~\ref{sec:conclusions} concludes the paper.

%% file: gecco-en4sr_related_work.tex
\section{Related work}
\label{sec:related}

One of the first works on NN in symbolic regression is Equation Learner (EQL), a multilayer feedforward network with sigmoid, sine, cosine, identity, and multiplication units \cite{martius2016}. The network is trained using the Adam stochastic gradient descent algorithm \cite{Kingma15} with a Lasso-like objective combining $L_2$ training loss and $L_1$ regularization. It uses a hybrid regularization strategy that starts with several update steps without regularization, followed by a regularization phase to enforce a sparse network structure to emerge. In the final phase, regularization is disabled, but the $L_0$ norm of the weights is still enforced, i.e., all weights close to 0 are set to 0.
In \cite{sahoo2018}, an extended version \eql was proposed, adding to the original EQL division units in the output layer.
Extension of EQL with other deep learning architectures and $L_{0.5}$ regularization were also proposed \cite{Kim2021}. 

OccamNet \cite{Costa2021} is another multilayer NN architecture partially inspired by EQL. It represents a probability distribution over functions, using a temperature-controlled connectivity scheme with skip connections similar to those in DenseNet \cite{Huang2017}. It uses a probabilistic interpretation of the softmax function by sampling sparse paths through a network to maximize sparsity. 
The Mathematical Operation Network (MathONet) proposed in \cite{Zhou2022} also uses an EQL-like NN architecture. The sparse subgraph of the NN is sought using a Bayesian learning approach that incorporates structural and nonstructural sparsity priors. 
The pruning algorithm for symbolic regression (PruneSymNet) proposed in \cite{wu2024prunesymnetsymbolicneuralnetwork} combines a greedy pruning algorithm with beam search to iteratively extract symbolic expressions from a differentiable neural network while maintaining its accuracy. This approach addresses the limitations of greedy algorithms in finding optimal solutions by generating multiple candidate expressions during pruning and selecting the one with the smallest loss.

A different class of NN-based SR approaches uses transformers such as GPT-2~\cite{radford19language}. They train the transformer model using a large amount of training data, where each sample is typically a tuple of the form (formula, data sampled from the formula). During inference, the transformer model constructs the formula based on a data set query. A discussion of the advantages and drawbacks of this approach is beyond the scope of this work. We refer interested readers to \cite{valipour21symbolicgpt, biggio21neural, dascoli22deep, Vastl2022, bendinelli2023, Bertschinger_srne_2024}.

Note that no prior knowledge is used in the approaches described above. When learning the model, these methods focus solely on minimizing the error computed on the training data. Consequently, they are likely to produce physically inconsistent results and do not generalize well to out-of-distribution samples.
A recent trend in the literature addresses this shortcoming by increasing the focus on integrating traditional physics-based modeling approaches with machine learning techniques \cite{Willard2020,vonRueden2023}. 
An interesting approach in this direction is the physics-informed neural network framework that trains deep neural networks to solve supervised learning tasks with relatively small amounts of training data and the underlying physics described by partial differential equations \cite{Raissi2019,LI2023105908}.

Combining SR with prior knowledge is still a relatively new research topic.
In \cite{Bladek2019}, counterexample-driven symbolic regression based on counterexample-driven genetic programming \cite{Krawiec2017} was introduced. It represents domain knowledge as a set of constraints expressed as logical formulas. A Satisfiability Modulo Theories solver is used to verify whether a candidate model meets the constraints. 
A similar approach called Logic Guided Genetic Algorithms (LGGA) was proposed in \cite{ashok2020logic}. Here, domain-specific knowledge is called auxiliary truths (AT), which are mathematical facts known a priori about the unknown function sought. ATs are represented as mathematical formulas. Candidate models are evaluated using a weighted sum of the classical training mean squared error and the truth error, which measures how much the model violates given ATs. Unlike CDSR, LGGA performs computationally light consistency checks via AT formula evaluations on the data set.

In \cite{Kubalik2020,Kubalik2021eswa}, a multi-objective SR method was proposed, which optimizes models with respect to training accuracy and the level of compliance with prior knowledge. Prior knowledge is defined as nonlinear inequality or equality constraints the system must obey. Synthetic constraint samples (i.e., samples not measured on the system) are generated for each constraint. Then, the constraint violation error measures how much the model violates the desired inequality or equality relations over the constraint samples.

The informed EQL (iEQL) proposed in \cite{werner2021} uses expert knowledge in the form of permitted or prohibited equation components and a domain-dependent structured sparsity prior. The authors demonstrated in experiments that iEQL learned interpretable and accurate models.
Shape-constrained symbolic regression was introduced in \cite{Kronberger2022evco,HAIDER2023,Martinek2024}. It allows for the inclusion of vague prior knowledge by measuring properties of the model's shape, such as monotonicity and convexity, using interval arithmetic. It has been shown that introducing shape constraints helps to find more realistic models.

%% file: gecco-en4sr_problem.tex
\section{Problem definition}
\label{sec:problem}

Consider a regression problem where a nonlinear model of an unknown function ${\phi: \mathbb{R}^n \rightarrow \mathbb{R}}$ is sought, with $n$ being the number of input variables. 
We seek a maximally sparse neural network model that minimizes the error observed on the validation data set and maximizes its compliance with the constraints imposed on the model, i.e., the desired properties of the model.
The constraint satisfaction measure is calculated on a set of samples explicitly generated for each type of constraint, as proposed in \cite{Kubalik2020,Kubalik2021eswa}.
The sparsity of the model is quantified by the number of active links and units in the neural network (see the definitions in Section~\ref{sec:master}).

Training, validation, and constraint data sets -- \Dt, \Dv, \Dc\ -- are used when learning and evaluating the generated models. 
The training data set contains samples of the form $\vecd_i = (\vecx_i, y_i)$, where $\vecx_i \in \mathbb{R}^n$ is sampled from the training domain \domTrain.
The validation data set contains samples of the same form as \Dt\ sampled from \domTrain such that $\Dv \cap \Dt =\varnothing$.
The constraint data adopt the constraint representation and evaluation scheme as proposed in \cite{Kubalik2020}. The set \Dc\ contains synthetic constraint samples generated for each constraint on which the constraint violation errors will be calculated.

%% file: gecco-en4sr_method.tex
\section{Method}
\label{sec:method}

The proposed algorithm generates candidate NN topologies using genetic operators and fine-tunes them via gradient-based backpropagation. Due to the inherent computational demands of evolutionary algorithms, it is not feasible to learn the parameters of each generated NN topology to full convergence. Instead, each NN topology is trained using only a short sequence of backpropagation iterations, while promising NN topologies are further refined in each generation.

Below, we describe the fundamental concepts of this method: the \textit{master topology}, a template from which candidate \textit{\sTops} are derived using crossover and mutation operators; a strategy to store and reuse the weights of the best-performing \sTops\ found so far to accelerate convergence to sparse \sTops; the genetic operators of crossover and mutation. Finally, we present a pseudocode of the method. The source code is publicly available at \url{https://github.com/jirkakubalik/EN4SR}.

\subsection{Master Topology}
\label{sec:master}

\begin{figure}[t]
\centerline{
\includegraphics[width=.925\columnwidth]{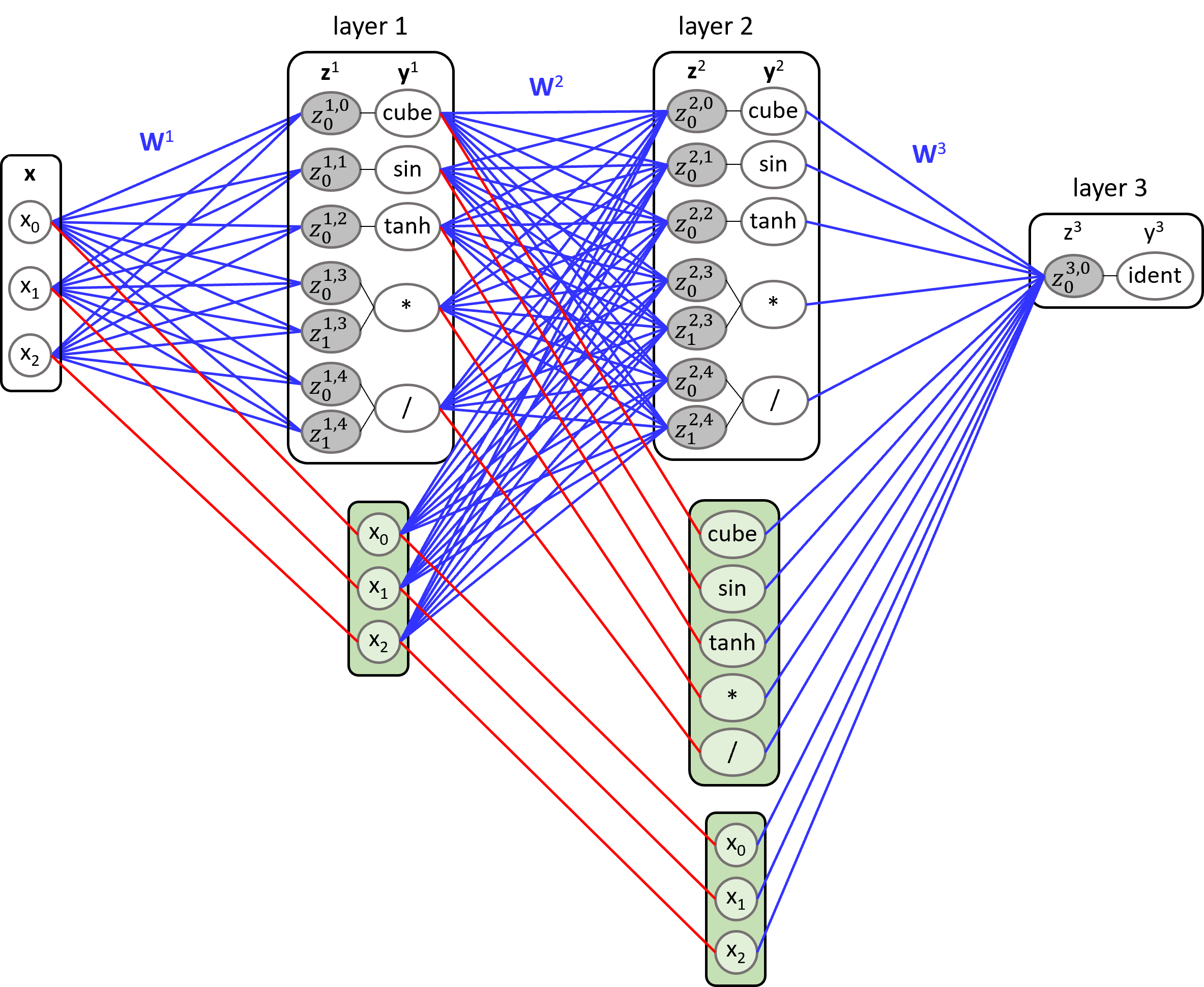}
}
\caption{Master topology with two hidden layers as proposed for \nnsr. 
The blue lines mark links with learnable weights. The red lines represent skip connections leading from the source units of layer $k-1$ to the copy units in layer $k$. For simplicity, this scheme does not show the bias links leading to every $z$-node.
}
\label{fig:master}
\end{figure}

The master topology represents the largest available network topology that can be used to encode the analytic expressions sought. It serves as a structural template for deriving \sTops. We adopt the multilayer feedforward NN architecture designed for \nnsr\ in \cite{kubalik_n4sr_2023}, see Figure~\ref{fig:master}. It is a heterogeneous NN with units implementing mathematical operators and elementary functions, such as \{+, --, *, /, sin, exp, etc.\}.

Each hidden layer contains \textit{learnable units} and \textit{copy units} (the term introduced in \cite{werner2021}). 
The learnable units are units whose input weights, denoted as \textit{learnable weights}, can be tuned by gradient-based learning (the links are shown in blue). All learnable weights of the master topology are collected in the set \vecWl.
The copy units in layer $k$ are just copying outputs of all units from the previous layer $k-1$, unit by unit. 
Each copy unit in layer $k$ has just a single input link, the \textit{skip connection} (the links shown in red), leading to its source unit in layer $k-1$. 
The skip connection values are set to 0 or 1 and are not subject to gradient-based learning.
With the skip connections, simple structures in shallow layers can be efficiently learned and used in subsequent layers. All skip connection weights are collected in \vecWs. The set of all master topology weights is $\vecW$.

Units represent \basis\ unary functions, e.g., $\sin$, $\cube$, $\tanh$, and binary functions (operators) such as multiplication * and division /.
The internal structure of the unit consists of an \textit{activation function} (denoted as `y' in Figure~\ref{fig:master}) and one or two \textit{z-nodes}, depending on the activation function's arity, serving as operands to the activation function. The $z$-nodes calculate an affine transformation of the output of its previous layer.
Each learnable unit $i$ in layer $j$ calculates its output as 
\begin{equation}
y^{j,i} = g(z^{j,i}_{0})\text{, for unary \basis\ function } g
\label{eq:y_unary}
\end{equation}
and
\begin{equation}
y^{j,i} = h(z^{j,i}_{0}, z^{j,i}_{1})\text{, for binary \basis\ function } h
\label{eq:y_binary}
\end{equation}
where $z^{j,i}_{\_}$ is calculated using the previous layer's output $\vecy^{j-1}$ as 
\begin{equation}
z^{j,i}_{\_} = \vecW^{j,i}_{l} \, \vecy^{j-1} + b^{j,i}
\label{eq:z}
\end{equation}
with learnable weights $\vecW^{j,i}_{l}$ and bias $b^{j,i}$. Likewise, in \nnsr, singularity units are allowed at any network layer. 

In the sequel, we use the following terminology:
\begin{itemize}
    \item \textit{Active link} -- a unit's input link\footnote{By the input link of a unit, we always mean the input links to the $z$-node(s) of that unit. An exception is the copy unit, which does not have a $z$-node and instead has a single input edge pointing directly to the corresponding unit in the previous layer.} whose weight $w \in \vecW$ has a positive absolute value. Links with a weight value equal to zero are considered \textit{inactive}.
    \item \textit{Active unit} -- a unit for which it holds that it has at least one active link and it is connected to the network output via a sequence of other active units. Thus, active units contribute to the network output. Units that do not contribute to the network output are considered \textit{inactive}.
    \item \textit{Enabled link} -- a unit's input link whose learnable weight $w \in \vecWl$ is updated through gradient-based learning. It can be both active and inactive.
    \item \textit{Disabled link} -- a unit's input link whose weight is set to zero and is not considered during gradient-based learning.
    \item $\thetaa$ -- a user-defined threshold that specifies the minimum absolute value a weight must have to be considered significant, see Section~\ref{sec:pseudocode}. 
    \item \textit{Enabled unit} -- a unit with at least one enabled link.
    \item \textit{Disabled unit} -- a unit whose all learnable weights are set to zero and cannot be updated through gradient-based learning.
\end{itemize}

\subsection{\sTopCap}
\label{sec:subtopology}

A \sTop\ is a NN derived from the master topology. Initially, it has all the units defined in the \mTopLow\ at its disposal.
When initializing a new \sTop, all units are active, all learnable units and links are enabled, and the learnable link values are initialized randomly. All skip connection values are set to 1.
Throughout the evolutionary process, only a subset of the units and their input links remain active.
Any learnable link can be switched from an enabled to a disabled state if its absolute weight value drops below the threshold $\thetaa$ at a certain point of the computation, see Section~\ref{sec:pseudocode}.
A unit can change its state from enabled to disabled and vice versa, either directly by the mutation operator, see Section~\ref{sec:genetic_operators}, or indirectly when all its input links become disabled.

The role of the multi-objective evolutionary algorithm described below is to create novel subtopologies by combining and modifying the current ones.
The standard stochastic gradient descent (SGD) algorithm, Adam~\cite{Kingma15}, is used to train the weights of the generated \sTops, using the following three loss functions adopted from \cite{kubalik_n4sr_2023}. Here, we only briefly characterize the individual loss functions and their components, for more details see \cite{kubalik_n4sr_2023}.
\begin{equation}
    \begin{split}
        \LI &= \Ltr + \Lsu \\
        \LII &= \Ltr + \Lsu + \Lcve \\
        \LIII &= \Ltr + \Lsu + \Lcve + \Lreg
    \end{split}
    \label{eq:loss}
\end{equation}
The loss functions are composed of the following terms:
\begin{itemize}
    \item Training loss, \Ltr\ -- the root-mean-square error (RMSE) calculated on the training data set \Dt.
    
    \item Singularity loss, \Lsu\ -- the root-mean-square \textit{singularity error} calculated on all available data samples, $\D = \Dt \cup \Dv \cup \Dc$, for all active singularity units. 
    The singularity error determines to what extent the singularity units correctly deal with the singularity values of their operands.
   
    \item Constraint loss, \Lcve\ -- this term accumulates the error of the model in terms of violating the prior knowledge. Typical constraints are the monotonicity of the function with respect to a given variable over a specific interval of its values, the symmetry of the model with respect to its arguments, etc.
    
    \item Regularization loss, \Lreg\ -- this term drives the learning process toward a sparse neural network representing a concise analytic expression. Here, we adopt the smoothed $L^{*}_{0.5}$ regularization, proposed in \cite{Kim2021}, applied to the set of active links of the \sTop.
\end{itemize}
Different loss functions and different numbers of learning steps are used in various stages of the algorithm, as shown in Algorithm~2.

\subsection{Weight memory}
\label{sec:transfer_learning}

An important component of this method is a strategy for managing and reusing the information acquired during the evolutionary process. This information is stored in $memory$ and reused by the crossover and mutation operators; see Section~\ref{sec:genetic_operators}. 

\subsubsection{Memory of $z$-node weights}
\label{sec:history}

Knowledge is stored in the \textit{memory} as the weights observed in the recent best-performing subtopologies. Memory is a structure containing a list of up to $s_{hist}$ records for every unit of the master topology, where $s_{hist}$ is a user-defined parameter. 
Each record is a tuple of $z$-node weights of the given learnable unit, i.e., a singleton with a single vector of weights and a pair of two weight vectors for unary and binary function units, respectively.
For copy units, the record is a scalar of either 0 or 1.
Note that the memory may contain an empty list of records for some unit if the unit has not been active in any subtopology used for memory updates. 

\subsubsection{Memory management}
\label{sec:history_management}

The memory is updated in each generation of the evolutionary algorithm. For the update, the well-performing and parsimonious subtopologies are selected from the current population of subtopologies using the following procedure:
\begin{enumerate}
    \item Extract the set $A$ of non-dominated subtopologies w.r.t. the validation RMSE, the singularity, and constraint losses. The regularization loss is omitted.
    \item Given the subtopologies in $A$, choose a set $B$ of solutions that have all its performance values below the median value of the subtopologies stored in $A$.
    \item Calculate the third quantile $q_3$ of the complexity of the solutions in $B$. Select a set $C$ consisting of subtopologies whose complexity is less than or equal to $q_3$.
\end{enumerate}
Steps 1) and 2) ensure the quality of selected solutions, while step 3) prevents the selection of overly complex solutions despite their good performance.

All active unit weights of the subtopologies in $C$ are used to update the memory in two steps. First, all records currently present in the memory originating from a subtopology that is dominated by at least one subtopology in $C$ are removed from the memory. Then, all subtopologies $s$ in $C$ are processed so that a new record for every active unit of the given subtopology $s$ is added to the memory if and only if the subtopology $s$ is not dominated by any subtopology currently represented in the memory by its records, and the size of the current list of records of the respective unit is less than $s_{hist}$.

\subsection{Genetic Operators}
\label{sec:genetic_operators}

This section describes the genetic operators designed to sample new \sTops\ with the use of the memory of weights.
When two \sTops\ $s_1$ and $s_2$ are crossed over, a new \sTop\ is generated in a layer-by-layer and a unit-by-unit manner while the child's $z$-node weights of the unit $u_{j,i}$ at $i^{th}$ position in $j^{th}$ layer are determined using the following structured rule where $p_l$, $p_i$, and $p_h$ are user-defined parameters
\begin{enumerate}
    \item If $rand() < p_{l}$, the \sTop\ $s_1$ is considered the lead; otherwise the $s_2$ is the lead.
    The lead \sTop\ determines how the weights will be generated for the given child unit.
    \item If $rand() < p_{i}$ and the unit $u_{j,i}$ is active in the lead \sTop\ then the weights of the lead's unit $u_{j,i}$ will be inherited.
    Otherwise, the weights will be taken from the memory with the probability $p_{h}$, or new weights will be randomly generated with the probability $1-p_{h}$. 
\end{enumerate}

The following two types of mutation operators are used to alter \sTops:
\begin{itemize}
    \item Unit status mutation -- this operator randomly selects one of the learnable units, which can be either enabled or disabled, and swaps its status. If a disabled unit turns to an enabled one, with probability $p_h$, its weights are taken randomly from the memory; otherwise, new weights are randomly generated. If an enabled unit turns to a disabled one, all its input weights are set to zero.
    \item Unit weights mutation -- this operator randomly selects one of the active units and changes its $z$-node weights to either the weights taken from the memory (with probability $p_h$) or the new randomly generated ones.
\end{itemize}

The role of the memory of weights is to accelerate gradient learning of the newly generated subtopologies and to speed up the overall convergence to the final sparse ones.

\subsection{Pseudo-code}
\label{sec:pseudocode}

The pseudo-code of the entire method is shown in Algorithm~\ref{alg:en4sr}. It starts by generating a population of random \sTops, initializing the weight memory, and the best-so-far set of non-dominated solutions based on the initial population. Non-dominated sorting is used that considers three \sTop performance criteria -- validation RMSE, constraint RMSE, and the number of active links.
Weights of the \sTops\ are learned for $N_t$ steps using \LI loss function.

The computation then proceeds in two nested loops: the outer loop represents stages, and the inner loop iterates over generations within each stage. Each stage begins with a perturbation of all \sTops\ in the current population (line 5). The perturbation means that all units are enabled, and their weights are assigned values the same way as in the mutation operator. This operator prevents irreversible stagnation in a suboptimal \sTop. After the \sTop\ is perturbed, its weights are trained for $N_t$ steps using the \LI loss function. A set of best stage \sTops, $stageBest$, is initialized using the perturbed population.

The body of the stage starts with generating new \sTops\ using crossover and mutation operators (line 9). Newly created \sTops\ undergo only a few gradient learning steps ($N_n$ is typically an order of units) with \LI loss function. In this way, many candidate \sTops\ can be generated. Out of the set of new \sTops, the best ones are chosen for the $interPop$ using the non-dominated sorting. Then, all \sTops\ in $interpop$ undergo gradient learning for a larger number of steps ($N_t$) using \LIII loss function. Moreover, the non-dominated \sTops\ of the current population are tuned for $N_f$ steps using \LII loss function. This means that the longer a \sTop\ remains among the non-dominated solutions in the population, the more gradient learning iterations it receives (in short batches). 
Next, the fine-tuned non-dominated solutions plus the solutions from the $interpop$ are merged with the current population, resulting in the new $pop$ for the next iteration of the stage. Finally, the set of the best stage \sTops\ is updated using the new population, and the memory of weights is updated based on the $stageBest$.
Once the stage finishes, the set of overall best \sTops\ is updated using the $stageBest$ solutions.

In the end, the algorithm returns the best \sTop, which is the least complex one that has all its performance criteria values below the median of the $overallBest$.

\LinesNumbered
\DontPrintSemicolon
\begin{algorithm}[t]
\fontsize{8.3}{9.6}\selectfont
\KwIn{Master $\dots$ master topology}
\myKwIn{$POPSIZE$ $\dots$ population size}
\myKwIn{$R$, $G_{R}$ $\dots$ number of stages, stage generations}
\myKwIn{$N_n$, $N_t$, $N_f$ $\dots$ newborn, tuning, fine-tuning backprop iterations}
\vspace{1mm}
\KwOut{best subtopology evolved}
\nonl\hrulefill\\
   $pop$.\init{\LI}; \;\nllabel{algXXX}
   $memory$.\init{$pop$}; $overallBest$.\init{$pop$}\;\nllabel{algXXX}
   r $\leftarrow$ 0 \;\nllabel{algXXX}
   \While{$r \leq$ R} {\nllabel{algXXX}
       $pop$.\perturb{\LI}\;\nllabel{algXXX}
       $stageBest$.\init{$pop$}\;\nllabel{algXXX}
       g $\leftarrow$ 0 \;\nllabel{algXXX}
       \While{$g \leq$ G} {\nllabel{algXXX}
          $interPop \leftarrow pop$.\runnsga{$N_n$, \LI} \;\nllabel{alg9}
          $interPop$.\runsgd{$N_t$, \LIII} \;\nllabel{algXXX}
          $nondominated \leftarrow pop$.\getnondom{} \;\nllabel{algXXX}
          $nondominated$.\runsgd{$N_f$, \LII} \;\nllabel{algXXX}
          $pop$.\merge{interPop, nondominated} \;\nllabel{algXXX}
          $stageBest$.\update{$pop$} \;\nllabel{algXXX}
          $memory$.\update{$stageBest$} \;\nllabel{algXXX}
          g $\leftarrow g+1$ \;\nllabel{algXXX}
       }
       $overallBest$.\update{$stageBest$} \;\nllabel{algXXX}
       r $\leftarrow r+1$ \;\nllabel{algXXX}
    }
\KwRet{$overallBest$.\selectbest{}} \;\nllabel{algXXX}
\caption{EN4SR}
\label{alg:en4sr}
\end{algorithm}

%% file: gecco-en4sr_experiments.tex
\section{Experiments}
\label{sec:experiments}

\subsection{Algorithms compared}
\label{sec:algorithms_compared}

We compared \ennsr\ with the NN-based algorithms \nnsrACYE\ and \eql, and a multi-objective GP-based algorithm using a local search procedure to estimate the coefficients of the evolved linear models \sngp, proposed in \cite{Kubalik2020}. Note that \nnsrACYE\ is the best-performing variant of the \nnsr\ algorithm \cite{kubalik_n4sr_2023}. 
We also test the \ennsr\ variant that does not use the weight memory, denoted as \ennsrBase.
Only \ennsr, \ennsrBase, \nnsrACYE, and \sngp\ can use prior knowledge about the model sought. 
The three algorithms compared were used with the same parameter setting as in \cite{kubalik_n4sr_2023}.

\subsection{Test problems}

Three problems, \resistors, \magic, and \magman\ adopted from \cite{kubalik_n4sr_2023} were used for proof-of-concept experiments. In addition, we applied the method to a real-world problem of dynamic system identification of a quadcopter. Below, we briefly summarize the problem specifics and data related to the first three problems; details can be found in \cite{kubalik_n4sr_2023}.\\
\begin{figure*}[h!]
\centerline{
    \subfigure[\resistors]{\includegraphics[width=.33\textwidth]{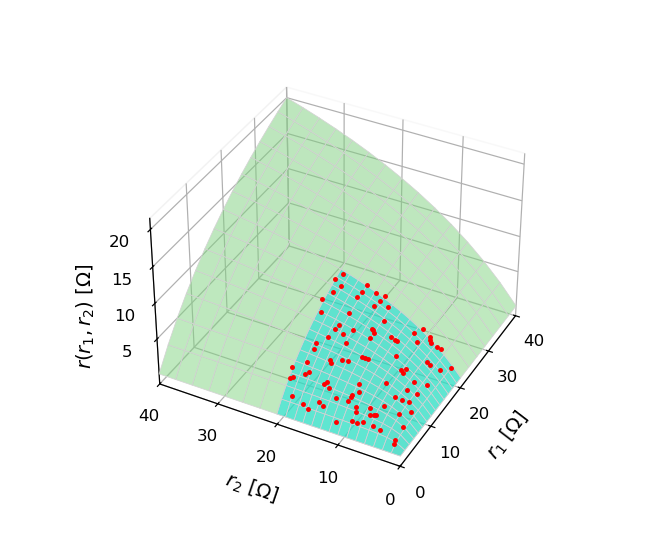}\label{fig:data_a}}
    \subfigure[\magic]{\includegraphics[width=.33\textwidth]{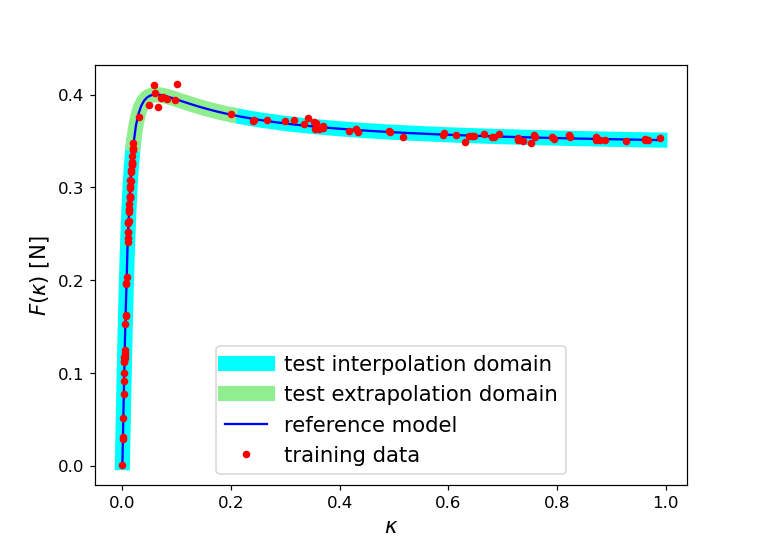}\label{fig:data_b}}
    \subfigure[\magman]{\includegraphics[width=.33\textwidth]{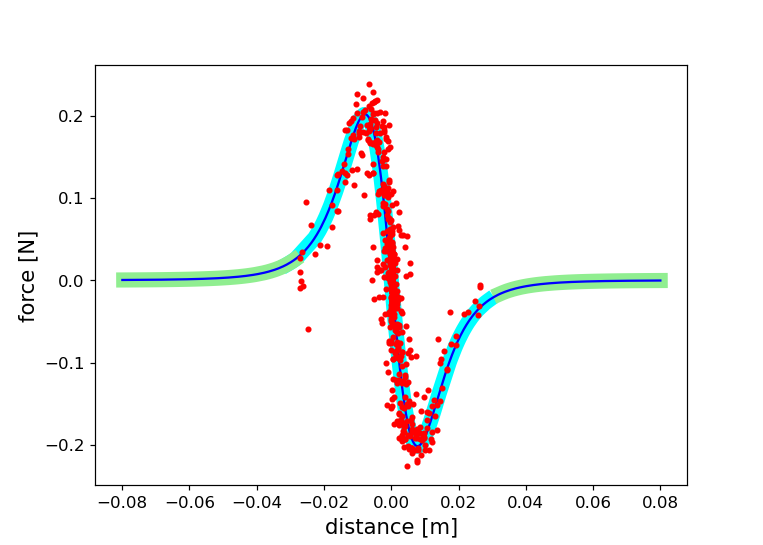}\label{fig:data_c}}
}
\caption{Reference models, training data $D_t$, interpolation domain $\mathbb{D}_i$, and extrapolation domain $\mathbb{D}_e$ of the three test problems.}
\label{fig:domains_data}
\end{figure*}

\noindent{\bf Resistors.} This problem was originally proposed in \cite{Bladek2019}. The task is to find a model of the equivalent resistance of two resistors in parallel, $r=r_1 r_2/(r_1+r_2)$, given noisy data \Dt\ (400 samples) and \Dv\ (100 samples) sampled uniformly from $\domTrain=[0.0001, 20]$\,$\Omega$ plus additional constraint samples \Dc\ representing prior knowledge: symmetry with respect to the arguments ($f(r_1,r_2) = f(r_2,r_1)$), diagonal property ($r_1=r_2 \Longrightarrow f(r_1,r_2)=\frac{r_1}{2}$), and upper-bound property ($f(r_1,r_2)\leq r_1$, $f(r_1,r_2)\leq r_2$). The \domTrain interval also serves as the interpolation interval \domInt to test the interpolation performance of generated models. The extrapolation performance of the models is tested on data sampled from the extrapolation interval $\domExt=[20.0001, 40]$, see Figure~\ref{fig:data_a}.\\
    
\noindent{\bf Magic.} This problem considers the tire-road interaction model, where the longitudinal force $F(\kappa)$ is the following function of the wheel slip $\kappa$ (aka the `magic' formula): 
$F(\kappa) = m\,g\,d\sin(c \arctan(b(1-e)\kappa + e\arctan(b\kappa)))$, where $b$, $c$, $d$ and $e$ are road surface-specific constants.
Here, we consider the \basemodel\ used in \cite{VERDIER2019} with $m=407.75$\,kg, $g=9.81$\,m\,$\cdot$\,s$^{-1}$, and the slip force parameters $(b, c, d, e) = (55.56, 1.35, 0.4, 0.52)$.
A data set of 110 samples was generated on \domTrain$=[0,1]$ using the reference model and divided into \Dt and \Dv in the ratio of 4:1. The data are intentionally sampled unevenly. The whole \domTrain interval is divided into $\domInt=[0,0.02] \cup [0.2, 0.99]$ and $\domExt=[0.03, 0.1]$, with well-represented \domInt and poorly-represented \domExt in \Dt, see Figure~\ref{fig:data_b}. 
The data deficiency is compensated by using prior knowledge of the three types: the function is zero for $\kappa=0$, it has a positive second derivative in the right part of \domInt, and it is concave down in \domExt.\\
    
\noindent{\bf Magman.} The magnetic manipulation system consists of an iron ball moving on a rail and an electromagnet placed at a fixed position under the rail. The goal is to find a nonlinear model of the magnetic force, $f(x)$, affecting the ball as a function of the horizontal distance, $x$, between the iron ball and the electromagnet, given a constant current $i$ through the coil.
The whole operating region of \magman\ is the interval $\domWhole=[-0.075,0.075]$\,m out of which only a small part $\domTrain=\domInt=[-0.027, 0.027]$\,m is covered by training and validation data sets $|\Dt|=400$ and $|\Dv|=201$. These data samples were measured on the real system.
The extrapolation data are sampled from the domain $\domExt=\domWhole \setminus \domInt$ using an empirical model $\tilde{f}(x)=-i c_1 x / (x^2+c_2)^3$ proposed in \cite{Hurak2012}, see Figure~\ref{fig:data_c}.
Five types of prior knowledge were defined: the function is odd; it is monotonically increasing on the intervals $[-0.075,-0.008]$\,m and $[0.008, 0.075]$\,m and monotonically decreasing on the interval $[-0.008,0.008]$\,m; it passes through the origin, i.e., $f(0)=0$; it quickly decays to zero as $x$ tends to infinity.

\subsection{Experiment set up}

We used the following two variants of the master topology, \mastera\ and \masterb, both with three hidden layers and a single identity unit ($\ident$) in the output layer. 
The $\ident$ unit calculates a weighted sum of its inputs to realize addition and subtraction.
The first two hidden layers in \mastera\ and \masterb\ contain four elementary functions $\{\sin, \tanh, \ident, *\}$ and \{$\sin, \arctan, \ident, *$\}, each with two copies. The third hidden layer contains, in addition to that, one division unit. Topology \mastera\ was used for \resistors\ and \magman, while \masterb\ was used for \magic\ problem.

We set the parameters of \nnsrACYE, \eql, and \sngp\ the same as in \cite{kubalik_n4sr_2023}.
Algorithm \ennsr\ was tested with the following parameter setting: $POPSIZE=10$, $R=3$, $G_R=20$, $N_n=10$, $N_t=100$, $N_f=50$.

Note that the maximum number of backprop iterations was set to 90000 for a fair comparison with \nnsr.
All tested methods used the same parameter setting on all test problems. No parameter tuning was carried out.

Thirty independent runs were performed with each method on every test problem. The best model is returned at the end of each run, yielding a set of thirty models on which we calculate the following performance measures:
\begin{itemize}
    \item complexity – model complexity defined as the number of active units and active units,
    \item $RMSE_{int+ext}$ -- RMSE calculated on test data sets sampling both the interpolation and extrapolation input domain of each test problem.
\end{itemize}
To assess the statistical significance of the observed differences in the $RMSE_{int+ext}$ performance of the compared algorithms, we use the Wilcoxon rank-sum test.

%% file: gecco-en4sr_results.tex
\subsection{Results}

Table~\ref{tab:results} presents the results obtained with the considered methods. 
We can see that \ennsr\ outperforms the NN- based \nnsrACYE\ on all problems in terms of $RMSE_{int+ext}$. This shows that the evolutionary component introduced in \ennsr\ helps to achieve better, though slightly larger, models.
\ennsr\ outperforms \eql\ on all problems that can be attributed to the fact that \eql\ does not use prior knowledge.
\ennsr\ also achieves significantly better results than the GP-based \sngp\ on \resistors, and performs equally well on the other two problems.
%
\begin{table}[h]
\caption{Results of \eql, \sngp, \nnsrACYE, \ennsrBase, and \ennsr\ on the \resistors, \magic, and \magman\ problem. The presented values are medians over 30 runs. The $p\text{-values}$ are calculated on the sets of $RMSE_{int+ext}$ values.} 
\label{tab:results}
\centering
\begin{tabular}{ l c c c }
\specialrule{1.5pt}{0pt}{0pt}  
Method & complexity & $RMSE_{int+ext}$ & $p\text{-value}$ \\
\specialrule{1.5pt}{0pt}{0pt}  
\multicolumn{4}{c}{{\bf\resistors}}  \\
\hline
\ennsr & 5\,/\,15.5 & $7.24 \cdot 10^{-3}$ & -- \\
\ennsrBase & 5\,/\,16 & $\mathbf{6.52 \cdot 10^{-3}}$ & 0.52 \\
\nnsrACYE & 3\,/\,10 & $4.7 \cdot 10^{-2}$ & $1.02 \cdot 10^{-5}$ \\
\hdashline[1.0pt/2pt]\\[-9pt]
\eql & 18\,/\,30 & $6.01$ & $9.4 \cdot 10^{-14}$ \\
\sngp & NA & $2.9 \cdot 10^{-2}$ & $2.7 \cdot 10^{-4}$ \\
\specialrule{1.5pt}{0pt}{0pt}  
\multicolumn{4}{c}{{\bf\magic}}  \\
\hline
\ennsr & 6\,/\,15.5 & $5.8 \cdot 10^{-3}$ & -- \\
\ennsrBase & 7.5\,/\,19 & $7.0 \cdot 10^{-3}$ & 0.92 \\
\nnsrACYE & 4\,/\,8.5 & $9.37 \cdot 10^{-2}$ & $1.9 \cdot 10^{-8}$ \\
\hdashline[1.0pt/2pt]\\[-9pt]
\eql & 17\,/\,29 & $5.26 \cdot 10^{-2}$ & $1.22 \cdot 10^{-9}$ \\
\sngp & NA & $\mathbf{2.01 \cdot 10^{-3}}$ & $0.21$ \\
\specialrule{1.5pt}{0pt}{0pt}  
\multicolumn{4}{c}{{\bf\magman}}  \\
\hline
\ennsr & 6\,/\,10.5 & $5.54 \cdot 10^{-2}$ & -- \\
\ennsrBase & 4\,/\,5.5 & $7.52 \cdot 10^{-2}$ & 0.015 \\
\nnsrACYE & 5\,/\,9 & $7.52 \cdot 10^{-2}$ & $7.58 \cdot 10^{-3}$ \\
\hdashline[1.0pt/2pt]\\[-9pt]
\eql & 18\,/\,30 & $5.86 \cdot 10^{-2}$ & 0.039 \\
\sngp & NA & $\mathbf{5.18 \cdot 10^{-2}}$ & 0.17 \\
\specialrule{1.5pt}{0pt}{0pt}  
\end{tabular}
\end{table}

\begin{figure}[t]
    \centering
    \subfigure[\ennsr]{\includegraphics[width=\columnwidth]{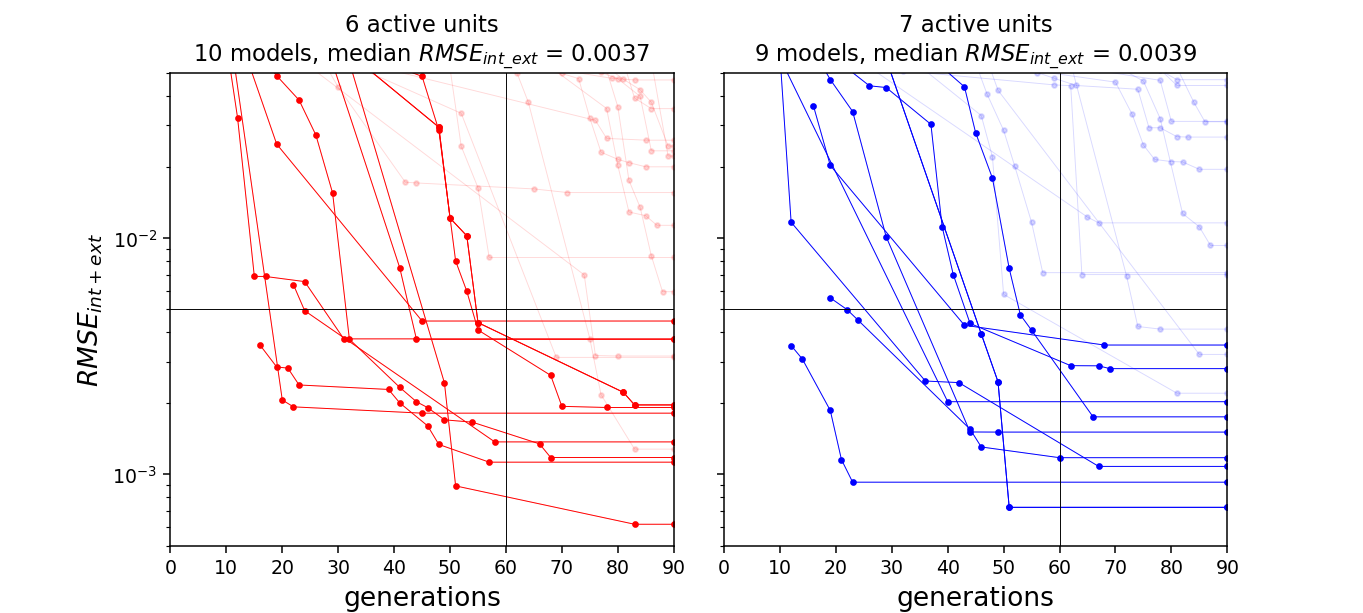}\label{fig:perf_magic_en4sr}}
    
    \vspace{-0.1cm}
    
    \subfigure[\ennsrBase]{\includegraphics[width=\columnwidth]{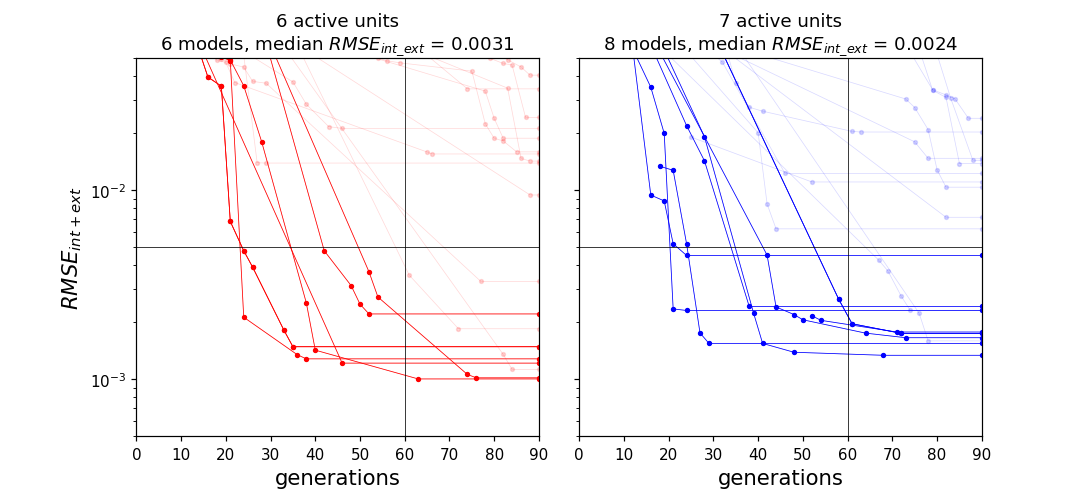}\label{fig:perf_magic_en4sr_base}}

    \caption{Convergence trajectories of models with 6 and 7 active units observed in 30 independent runs with \ennsr\ and \ennsrBase\ on \magic\ problem.}
    \label{fig:perf_resistors}
\end{figure}

Figures~\ref{fig:perf_magic_en4sr} and \ref{fig:perf_magic_en4sr_base} illustrate different convergence properties of \ennsr\ and \ennsrBase\ on \magic\ problem where the algorithms exhibit otherwise very similar performance ($p$-value\,=\,0.92).
The figures show convergence trajectories of models with 6 and 7 active units observed in 30 independent runs. 
For each run an evolution of the best $RMSE_{int+ext}$ value of a model with the given complexity is shown if such a model appears in the run (trajectories are plotted in light shades of red and blue color). 
Some of these trajectories are shown in saturated colors if a model of the given complexity with a $RMSE_{int+ext}$ value less than $5 \cdot 10^{-3}$ appeared in the run for the first time before generation 60. The generation number and the quality threshold were chosen to demonstrate the different behavior of the algorithms during the phase in which high-quality models begin to emerge.
We can see, though it is hard to quantify, a difference in the number and distribution of these small and high-quality models. 

\subsection{Dynamic model of Quadcopter\label{sec:exp_quadcopter}}

In this section, we use real-world data from the Parrot Bebop 2 quadcopter to demonstrate the method's performance in a dynamic system identification task. The measured variables are the translational velocities $v_x$, $v_y$, and $v_z$ in the fixed world frame (acquired in an indoor flying arena with the OptiTrack motion-capture system) and the body angles $\theta$, $\varphi$ and $\psi$, denoting pitch, roll, and yaw, respectively. The quadcopter was controlled by a model predictive controller (MPC) using the desired roll, pitch, yaw rate, and vertical velocity as control input, with the sampling period $T_s =0.05$~s. The data set contains 498 samples acquired in an experiment with the quadcopter following an 8-shape trajectory in the horizontal plane (at a constant altitude). The data collected were divided into the training set (360 samples), the validation set (90 samples), and the test set (48 samples).
The master topology had two hidden layers, each with one sin and cos unit, and two ident and multiplier units.

We use the proposed method to build a prediction model for the translational velocity $v_x$ in the form:
$$
 v_x(k+1) = g(v_x(k), \theta(k)),
$$
where $k = 0,1,2,\ldots$ denotes the discrete time instant and $g$ is the unknown function sought by symbolic regression.

\begin{figure}[htbp]
\centerline{\includegraphics[width=.45\textwidth]{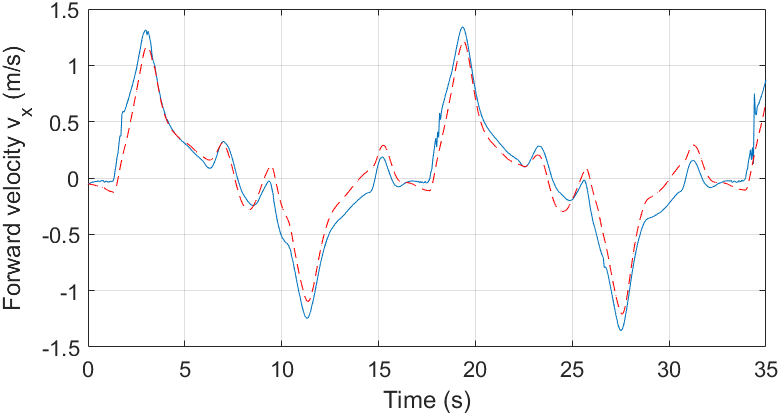}}
\caption{Quadcopter model simulation on unseen test data. Blue solid line: measured data, red dashed line: model.}
\label{fig:quadcopter-simulation}
\end{figure}

We generated 30 models, of which 24 turned out to be linear $v_x(k+1) =  0.985v_x(k) + 0.473\theta(k)$, where the coefficients were rounded to three decimal digits. The remaining six models contained a sine or cosine nonlinearity. The 30 models performed almost equally well in a test simulation run on a fresh data set that was not used in the training. The test data set contains 699 samples acquired in an experiment with the quadcopter following a square trajectory in the horizontal plane. Figure~\ref{fig:quadcopter-simulation} illustrates the performance of the above linear model. This recursive simulation of the model, $\hat{v}_x(k+1) = 0.985\hat{v}_x(k) + 0.473\theta(k)$, demonstrates the ability of the SR method to discover accurate predictive models of the quadcopter dynamics from data. The fact that the models do not contain an affine term (offset) represents the fact that the data were acquired indoors in the absence of wind. In an outdoor situation, the affine term would estimate the wind speed component in the respective direction.

%% file: gecco-en4sr_conclusions.tex
\section{Conclusions and future work}
\label{sec:conclusions}

We proposed a new neuro-evolutionary symbolic regression method, \ennsr, which combines an evolutionary search for a suitable neural network topology with gradient-descent learning of the network's coefficients. We experimentally evaluated the proposed method on three tested problems. We also demonstrated the method’s performance in a dynamic system identification task using real-world data.
With a similar computing budget, this approach outperforms other NN-based approaches. It is also computationally more efficient and less prone to getting stuck in local optima. This clearly indicates the positive impact of the evolutionary component implemented in the method.

Note that the complexity of evolved neural network topologies measured by the number of active units and links relates only approximately to the complexity of the final analytic expression reduced to its minimal form. In our future research, we will address the problem of complexity reduction throughout the learning process via symbolic simplification. We will also analyze the effect of the memory-based strategy for maintaining and reusing useful information acquired during the evolutionary process and investigate new possibilities to realize this feature. 

Since the proposed method is based on a multi-objective optimization approach, it inherently produces a set of non-dominated solutions. We have observed that the current final model selection strategy often overlooks models that perform notably better than the one ultimately selected. Therefore, we will also focus on designing a final model selection strategy that mitigates this undesired effect.

%% file: gecco-en4sr_00_main.bbl

\begin{thebibliography}{45}


\ifx \showCODEN    \undefined \def \showCODEN     #1{\unskip}     \fi
\ifx \showISBNx    \undefined \def \showISBNx     #1{\unskip}     \fi
\ifx \showISBNxiii \undefined \def \showISBNxiii  #1{\unskip}     \fi
\ifx \showISSN     \undefined \def \showISSN      #1{\unskip}     \fi
\ifx \showLCCN     \undefined \def \showLCCN      #1{\unskip}     \fi
\ifx \shownote     \undefined \def \shownote      #1{#1}          \fi
\ifx \showarticletitle \undefined \def \showarticletitle #1{#1}   \fi
\ifx \showURL      \undefined \def \showURL       {\relax}        \fi
\providecommand\bibfield[2]{#2}
\providecommand\bibinfo[2]{#2}
\providecommand\natexlab[1]{#1}
\providecommand\showeprint[2][]{arXiv:#2}

\bibitem[Arnaldo et~al\mbox{.}(2015)]%
        {Arnaldo2015}
\bibfield{author}{\bibinfo{person}{Ignacio Arnaldo}, \bibinfo{person}{Una-May O’Reilly}, {and} \bibinfo{person}{Kalyan Veeramachaneni}.} \bibinfo{year}{2015}\natexlab{}.
\newblock \showarticletitle{Building Predictive Models via Feature Synthesis}. In \bibinfo{booktitle}{\emph{Proceedings of the 2015 Annual Conference on Genetic and Evolutionary Computation}} (Madrid, Spain) \emph{(\bibinfo{series}{GECCO ’15})}. \bibinfo{publisher}{Association for Computing Machinery}, \bibinfo{address}{New York, NY, USA}, \bibinfo{pages}{983–990}.
\newblock
\showISBNx{9781450334723}
\href{https://doi.org/10.1145/2739480.2754693}{doi:\nolinkurl{10.1145/2739480.2754693}}


\bibitem[Ashok et~al\mbox{.}(2020)]%
        {ashok2020logic}
\bibfield{author}{\bibinfo{person}{Dhananjay Ashok}, \bibinfo{person}{Joseph Scott}, \bibinfo{person}{Sebastian Wetzel}, \bibinfo{person}{Maysum Panju}, {and} \bibinfo{person}{Vijay Ganesh}.} \bibinfo{year}{2020}\natexlab{}.
\newblock \bibinfo{title}{Logic Guided Genetic Algorithms}.
\newblock
\showeprint[arxiv]{2010.11328}~[cs.NE]


\bibitem[Bendinelli et~al\mbox{.}(2023)]%
        {bendinelli2023}
\bibfield{author}{\bibinfo{person}{Tommaso Bendinelli}, \bibinfo{person}{Luca Biggio}, {and} \bibinfo{person}{Pierre-Alexandre Kamienny}.} \bibinfo{year}{2023}\natexlab{}.
\newblock \bibinfo{title}{Controllable Neural Symbolic Regression}.
\newblock
\showeprint[arxiv]{2304.10336}~[cs.LG]
\urldef\tempurl%
\url{https://arxiv.org/abs/2304.10336}
\showURL{%
\tempurl}


\bibitem[Bertschinger et~al\mbox{.}(2024)]%
        {Bertschinger_srne_2024}
\bibfield{author}{\bibinfo{person}{Amanda Bertschinger}, \bibinfo{person}{James Bagrow}, {and} \bibinfo{person}{Joshua Bongard}.} \bibinfo{year}{2024}\natexlab{}.
\newblock \showarticletitle{Evolving Form and Function: Dual-Objective Optimization in Neural Symbolic Regression Networks}. \bibinfo{pages}{277--285}.
\newblock
\href{https://doi.org/10.1145/3638529.3654030}{doi:\nolinkurl{10.1145/3638529.3654030}}


\bibitem[Biggio et~al\mbox{.}(2021)]%
        {biggio21neural}
\bibfield{author}{\bibinfo{person}{Luca Biggio}, \bibinfo{person}{Tommaso Bendinelli}, \bibinfo{person}{Alexander Neitz}, \bibinfo{person}{Aur{\'{e}}lien Lucchi}, {and} \bibinfo{person}{Giambattista Parascandolo}.} \bibinfo{year}{2021}\natexlab{}.
\newblock \showarticletitle{Neural Symbolic Regression that Scales}.
\newblock \bibinfo{journal}{\emph{CoRR}}  \bibinfo{volume}{abs/2106.06427} (\bibinfo{year}{2021}).
\newblock
\showeprint[arXiv]{2106.06427}
\urldef\tempurl%
\url{https://arxiv.org/abs/2106.06427}
\showURL{%
\tempurl}


\bibitem[Bladek and Krawiec(2019)]%
        {Bladek2019}
\bibfield{author}{\bibinfo{person}{Iwo Bladek} {and} \bibinfo{person}{Krzysztof Krawiec}.} \bibinfo{year}{2019}\natexlab{}.
\newblock \showarticletitle{Solving Symbolic Regression Problems with Formal Constraints}. In \bibinfo{booktitle}{\emph{Proceedings of the Genetic and Evolutionary Computation Conference}} (Prague, Czech Republic) \emph{(\bibinfo{series}{GECCO '19})}. \bibinfo{publisher}{ACM}, \bibinfo{address}{New York, NY, USA}, \bibinfo{pages}{977--984}.
\newblock
\showISBNx{978-1-4503-6111-8}
\href{https://doi.org/10.1145/3321707.3321743}{doi:\nolinkurl{10.1145/3321707.3321743}}


\bibitem[Burlacu et~al\mbox{.}(2020)]%
        {Burlacu2020OperonCA}
\bibfield{author}{\bibinfo{person}{Bogdan Burlacu}, \bibinfo{person}{Gabriel Kronberger}, {and} \bibinfo{person}{Michael Kommenda}.} \bibinfo{year}{2020}\natexlab{}.
\newblock \showarticletitle{Operon C++: an efficient genetic programming framework for symbolic regression}.
\newblock \bibinfo{journal}{\emph{Proceedings of the 2020 Genetic and Evolutionary Computation Conference Companion}} (\bibinfo{year}{2020}).
\newblock
\urldef\tempurl%
\url{https://api.semanticscholar.org/CorpusID:220409306}
\showURL{%
\tempurl}


\bibitem[Costa et~al\mbox{.}(2021)]%
        {Costa2021}
\bibfield{author}{\bibinfo{person}{Allan Costa}, \bibinfo{person}{Rumen Dangovski}, \bibinfo{person}{Owen Dugan}, \bibinfo{person}{Samuel Kim}, \bibinfo{person}{Pawan Goyal}, \bibinfo{person}{Marin Soljačić}, {and} \bibinfo{person}{Joseph Jacobson}.} \bibinfo{year}{2021}\natexlab{}.
\newblock \bibinfo{title}{Fast Neural Models for Symbolic Regression at Scale}.
\newblock
\href{https://doi.org/10.48550/ARXIV.2007.10784}{doi:\nolinkurl{10.48550/ARXIV.2007.10784}}


\bibitem[d'Ascoli et~al\mbox{.}(2022)]%
        {dascoli22deep}
\bibfield{author}{\bibinfo{person}{St{\'{e}}phane d'Ascoli}, \bibinfo{person}{Pierre{-}Alexandre Kamienny}, \bibinfo{person}{Guillaume Lample}, {and} \bibinfo{person}{Fran{\c{c}}ois Charton}.} \bibinfo{year}{2022}\natexlab{}.
\newblock \showarticletitle{Deep Symbolic Regression for Recurrent Sequences}.
\newblock \bibinfo{journal}{\emph{CoRR}}  \bibinfo{volume}{abs/2201.04600} (\bibinfo{year}{2022}).
\newblock
\showeprint[arXiv]{2201.04600}
\urldef\tempurl%
\url{https://arxiv.org/abs/2201.04600}
\showURL{%
\tempurl}


\bibitem[Derner et~al\mbox{.}(2020)]%
        {Derner2020}
\bibfield{author}{\bibinfo{person}{Erik Derner}, \bibinfo{person}{Jir{\'{\i}} Kubal{\'{\i}}k}, \bibinfo{person}{Nicola Ancona}, {and} \bibinfo{person}{Robert Babuska}.} \bibinfo{year}{2020}\natexlab{}.
\newblock \showarticletitle{Constructing parsimonious analytic models for dynamic systems via symbolic regression}.
\newblock \bibinfo{journal}{\emph{Appl. Soft Comput.}}  \bibinfo{volume}{94} (\bibinfo{year}{2020}), \bibinfo{pages}{106432}.
\newblock
\href{https://doi.org/10.1016/j.asoc.2020.106432}{doi:\nolinkurl{10.1016/j.asoc.2020.106432}}


\bibitem[Derner et~al\mbox{.}(2021)]%
        {Derner21}
\bibfield{author}{\bibinfo{person}{Erik Derner}, \bibinfo{person}{Jir{\'{\i}} Kubal{\'{\i}}k}, {and} \bibinfo{person}{Robert Babuska}.} \bibinfo{year}{2021}\natexlab{}.
\newblock \showarticletitle{Selecting Informative Data Samples for Model Learning Through Symbolic Regression}.
\newblock \bibinfo{journal}{\emph{{IEEE} Access}}  \bibinfo{volume}{9} (\bibinfo{year}{2021}), \bibinfo{pages}{14148--14158}.
\newblock
\href{https://doi.org/10.1109/ACCESS.2021.3052130}{doi:\nolinkurl{10.1109/ACCESS.2021.3052130}}


\bibitem[Haider et~al\mbox{.}(2023)]%
        {HAIDER2023}
\bibfield{author}{\bibinfo{person}{C. Haider}, \bibinfo{person}{F.O. {de Franca}}, \bibinfo{person}{B. Burlacu}, {and} \bibinfo{person}{G. Kronberger}.} \bibinfo{year}{2023}\natexlab{}.
\newblock \showarticletitle{Shape-constrained multi-objective genetic programming for symbolic regression}.
\newblock \bibinfo{journal}{\emph{Applied Soft Computing}}  \bibinfo{volume}{132} (\bibinfo{year}{2023}), \bibinfo{pages}{109855}.
\newblock
\showISSN{1568-4946}
\href{https://doi.org/10.1016/j.asoc.2022.109855}{doi:\nolinkurl{10.1016/j.asoc.2022.109855}}


\bibitem[Huang et~al\mbox{.}(2017)]%
        {Huang2017}
\bibfield{author}{\bibinfo{person}{Gao Huang}, \bibinfo{person}{Zhuang Liu}, \bibinfo{person}{Laurens Van Der~Maaten}, {and} \bibinfo{person}{Kilian~Q. Weinberger}.} \bibinfo{year}{2017}\natexlab{}.
\newblock \showarticletitle{Densely Connected Convolutional Networks}. In \bibinfo{booktitle}{\emph{2017 IEEE Conference on Computer Vision and Pattern Recognition (CVPR)}}. \bibinfo{pages}{2261--2269}.
\newblock
\href{https://doi.org/10.1109/CVPR.2017.243}{doi:\nolinkurl{10.1109/CVPR.2017.243}}


\bibitem[Hurak and Zemanek(2012)]%
        {Hurak2012}
\bibfield{author}{\bibinfo{person}{Zdenek Hurak} {and} \bibinfo{person}{Jiri Zemanek}.} \bibinfo{year}{2012}\natexlab{}.
\newblock \showarticletitle{Feedback linearization approach to distributed feedback manipulation}. In \bibinfo{booktitle}{\emph{American control conference}}. \bibinfo{address}{Montreal, Canada}, \bibinfo{pages}{991--996}.
\newblock
\showISBNx{9781457710964}


\bibitem[Kim et~al\mbox{.}(2021)]%
        {Kim2021}
\bibfield{author}{\bibinfo{person}{Samuel Kim}, \bibinfo{person}{Peter~Y. Lu}, \bibinfo{person}{Srijon Mukherjee}, \bibinfo{person}{Michael Gilbert}, \bibinfo{person}{Li Jing}, \bibinfo{person}{Vladimir Ceperic}, {and} \bibinfo{person}{Marin Soljacic}.} \bibinfo{year}{2021}\natexlab{}.
\newblock \showarticletitle{Integration of Neural Network-Based Symbolic Regression in Deep Learning for Scientific Discovery}.
\newblock \bibinfo{journal}{\emph{{IEEE} Transactions on Neural Networks and Learning Systems}} \bibinfo{volume}{32}, \bibinfo{number}{9} (\bibinfo{date}{sep} \bibinfo{year}{2021}), \bibinfo{pages}{4166--4177}.
\newblock
\href{https://doi.org/10.1109/tnnls.2020.3017010}{doi:\nolinkurl{10.1109/tnnls.2020.3017010}}


\bibitem[Kingma and Ba(2015)]%
        {Kingma15}
\bibfield{author}{\bibinfo{person}{Diederick~P Kingma} {and} \bibinfo{person}{Jimmy Ba}.} \bibinfo{year}{2015}\natexlab{}.
\newblock \showarticletitle{Adam: A method for stochastic optimization}. In \bibinfo{booktitle}{\emph{International Conference on Learning Representations (ICLR)}}.
\newblock


\bibitem[Kommenda et~al\mbox{.}(2020)]%
        {Kommenda2020}
\bibfield{author}{\bibinfo{person}{Michael Kommenda}, \bibinfo{person}{Bogdan Burlacu}, \bibinfo{person}{Gabriel Kronberger}, {and} \bibinfo{person}{Michael Affenzeller}.} \bibinfo{year}{2020}\natexlab{}.
\newblock \showarticletitle{Parameter Identification for Symbolic Regression Using Nonlinear Least Squares}.
\newblock \bibinfo{journal}{\emph{Genetic Programming and Evolvable Machines}} \bibinfo{volume}{21}, \bibinfo{number}{3} (\bibinfo{date}{sep} \bibinfo{year}{2020}), \bibinfo{pages}{471–501}.
\newblock
\showISSN{1389-2576}
\href{https://doi.org/10.1007/s10710-019-09371-3}{doi:\nolinkurl{10.1007/s10710-019-09371-3}}


\bibitem[Koza(1992)]%
        {Koza1992}
\bibfield{author}{\bibinfo{person}{John~R. Koza}.} \bibinfo{year}{1992}\natexlab{}.
\newblock \bibinfo{booktitle}{\emph{Genetic Programming: On the Programming of Computers by Means of Natural Selection}}.
\newblock \bibinfo{publisher}{MIT Press}, \bibinfo{address}{Cambridge, MA, USA}.
\newblock
\showISBNx{0262111705}


\bibitem[Krawiec et~al\mbox{.}(2017)]%
        {Krawiec2017}
\bibfield{author}{\bibinfo{person}{Krzysztof Krawiec}, \bibinfo{person}{Iwo Bladek}, {and} \bibinfo{person}{Jerry Swan}.} \bibinfo{year}{2017}\natexlab{}.
\newblock \showarticletitle{Counterexample-driven Genetic Programming}. In \bibinfo{booktitle}{\emph{Proceedings of the Genetic and Evolutionary Computation Conference}} (Berlin, Germany) \emph{(\bibinfo{series}{GECCO '17})}. \bibinfo{publisher}{ACM}, \bibinfo{address}{New York, NY, USA}, \bibinfo{pages}{953--960}.
\newblock
\showISBNx{978-1-4503-4920-8}
\href{https://doi.org/10.1145/3071178.3071224}{doi:\nolinkurl{10.1145/3071178.3071224}}


\bibitem[Kronberger et~al\mbox{.}(2022)]%
        {Kronberger2022evco}
\bibfield{author}{\bibinfo{person}{G. Kronberger}, \bibinfo{person}{F.~O. de Franca}, \bibinfo{person}{B. Burlacu}, \bibinfo{person}{C. Haider}, {and} \bibinfo{person}{M. Kommenda}.} \bibinfo{year}{2022}\natexlab{}.
\newblock \showarticletitle{{Shape-Constrained Symbolic Regression—Improving Extrapolation with Prior Knowledge}}.
\newblock \bibinfo{journal}{\emph{Evolutionary Computation}} \bibinfo{volume}{30}, \bibinfo{number}{1} (\bibinfo{date}{03} \bibinfo{year}{2022}), \bibinfo{pages}{75--98}.
\newblock
\showISSN{1063-6560}
\href{https://doi.org/10.1162/evco_a_00294}{doi:\nolinkurl{10.1162/evco_a_00294}}


\bibitem[Kubal{\'{\i}}k et~al\mbox{.}(2021)]%
        {Kubalik2021eswa}
\bibfield{author}{\bibinfo{person}{Jir{\'{\i}} Kubal{\'{\i}}k}, \bibinfo{person}{Erik Derner}, {and} \bibinfo{person}{Robert Babuska}.} \bibinfo{year}{2021}\natexlab{}.
\newblock \showarticletitle{Multi-objective symbolic regression for physics-aware dynamic modeling}.
\newblock \bibinfo{journal}{\emph{Expert Syst. Appl.}}  \bibinfo{volume}{182} (\bibinfo{year}{2021}), \bibinfo{pages}{115210}.
\newblock
\href{https://doi.org/10.1016/j.eswa.2021.115210}{doi:\nolinkurl{10.1016/j.eswa.2021.115210}}


\bibitem[Kubal\'{\i}k et~al\mbox{.}(2020)]%
        {Kubalik2020}
\bibfield{author}{\bibinfo{person}{Ji\v{r}\'{\i} Kubal\'{\i}k}, \bibinfo{person}{Erik Derner}, {and} \bibinfo{person}{Robert Babu\v{s}ka}.} \bibinfo{year}{2020}\natexlab{}.
\newblock \showarticletitle{Symbolic Regression Driven by Training Data and Prior Knowledge}. In \bibinfo{booktitle}{\emph{Proceedings of the 2020 Genetic and Evolutionary Computation Conference}} (Canc\'{u}n, Mexico) \emph{(\bibinfo{series}{GECCO '20})}. \bibinfo{publisher}{Association for Computing Machinery}, \bibinfo{address}{New York, NY, USA}, \bibinfo{pages}{958–966}.
\newblock
\showISBNx{9781450371285}
\href{https://doi.org/10.1145/3377930.3390152}{doi:\nolinkurl{10.1145/3377930.3390152}}


\bibitem[Kubalík et~al\mbox{.}(2023)]%
        {kubalik_n4sr_2023}
\bibfield{author}{\bibinfo{person}{Jiří Kubalík}, \bibinfo{person}{Erik Derner}, {and} \bibinfo{person}{Robert Babuška}.} \bibinfo{year}{2023}\natexlab{}.
\newblock \showarticletitle{Toward Physically Plausible Data-Driven Models: A Novel Neural Network Approach to Symbolic Regression}.
\newblock \bibinfo{journal}{\emph{IEEE Access}}  \bibinfo{volume}{11} (\bibinfo{year}{2023}), \bibinfo{pages}{61481--61501}.
\newblock
\href{https://doi.org/10.1109/ACCESS.2023.3287397}{doi:\nolinkurl{10.1109/ACCESS.2023.3287397}}


\bibitem[Li et~al\mbox{.}(2023)]%
        {LI2023105908}
\bibfield{author}{\bibinfo{person}{Shilin Li}, \bibinfo{person}{Gang Wang}, \bibinfo{person}{Yuelan Di}, \bibinfo{person}{Liping Wang}, \bibinfo{person}{Haidou Wang}, {and} \bibinfo{person}{Qingjun Zhou}.} \bibinfo{year}{2023}\natexlab{}.
\newblock \showarticletitle{A physics-informed neural network framework to predict 3D temperature field without labeled data in process of laser metal deposition}.
\newblock \bibinfo{journal}{\emph{Engineering Applications of Artificial Intelligence}}  \bibinfo{volume}{120} (\bibinfo{year}{2023}), \bibinfo{pages}{105908}.
\newblock
\showISSN{0952-1976}
\href{https://doi.org/10.1016/j.engappai.2023.105908}{doi:\nolinkurl{10.1016/j.engappai.2023.105908}}


\bibitem[Martinek et~al\mbox{.}(2024)]%
        {Martinek2024}
\bibfield{author}{\bibinfo{person}{Viktor Martinek}, \bibinfo{person}{Julia Reuter}, \bibinfo{person}{Ophelia Frotscher}, \bibinfo{person}{Sanaz Mostaghim}, \bibinfo{person}{Markus Richter}, {and} \bibinfo{person}{Roland Herzog}.} \bibinfo{year}{2024}\natexlab{}.
\newblock \showarticletitle{Shape Constraints in Symbolic Regression using Penalized Least Squares}.
\newblock \bibinfo{journal}{\emph{CoRR}}  \bibinfo{volume}{abs/2405.20800} (\bibinfo{year}{2024}).
\newblock
\href{https://doi.org/10.48550/ARXIV.2405.20800}{doi:\nolinkurl{10.48550/ARXIV.2405.20800}}
\showeprint[arXiv]{2405.20800}


\bibitem[Martius and Lampert(2016)]%
        {martius2016}
\bibfield{author}{\bibinfo{person}{Georg Martius} {and} \bibinfo{person}{Christoph~H. Lampert}.} \bibinfo{year}{2016}\natexlab{}.
\newblock \showarticletitle{Extrapolation and learning equations}.
\newblock \bibinfo{journal}{\emph{CoRR}}  \bibinfo{volume}{abs/1610.02995} (\bibinfo{year}{2016}).
\newblock
\showeprint[arXiv]{1610.02995}
\urldef\tempurl%
\url{http://arxiv.org/abs/1610.02995}
\showURL{%
\tempurl}


\bibitem[Radford et~al\mbox{.}(2019)]%
        {radford19language}
\bibfield{author}{\bibinfo{person}{Alec Radford}, \bibinfo{person}{Jeff Wu}, \bibinfo{person}{Rewon Child}, \bibinfo{person}{David Luan}, \bibinfo{person}{Dario Amodei}, {and} \bibinfo{person}{Ilya Sutskever}.} \bibinfo{year}{2019}\natexlab{}.
\newblock \bibinfo{title}{Language Models are Unsupervised Multitask Learners}.
\newblock


\bibitem[Raissi et~al\mbox{.}(2019)]%
        {Raissi2019}
\bibfield{author}{\bibinfo{person}{M. Raissi}, \bibinfo{person}{P. Perdikaris}, {and} \bibinfo{person}{G.E. Karniadakis}.} \bibinfo{year}{2019}\natexlab{}.
\newblock \showarticletitle{Physics-informed neural networks: A deep learning framework for solving forward and inverse problems involving nonlinear partial differential equations}.
\newblock \bibinfo{journal}{\emph{J. Comput. Phys.}}  \bibinfo{volume}{378} (\bibinfo{year}{2019}), \bibinfo{pages}{686--707}.
\newblock
\showISSN{0021-9991}
\href{https://doi.org/10.1016/j.jcp.2018.10.045}{doi:\nolinkurl{10.1016/j.jcp.2018.10.045}}


\bibitem[Richmond-Navarro et~al\mbox{.}(2017)]%
        {Richmond2017}
\bibfield{author}{\bibinfo{person}{Gustavo Richmond-Navarro}, \bibinfo{person}{Williams~R. Calderón-Muñoz}, \bibinfo{person}{Richard LeBoeuf}, {and} \bibinfo{person}{Pablo Castillo}.} \bibinfo{year}{2017}\natexlab{}.
\newblock \showarticletitle{A Magnus Wind Turbine Power Model Based on Direct Solutions Using the Blade Element Momentum Theory and Symbolic Regression}.
\newblock \bibinfo{journal}{\emph{IEEE Transactions on Sustainable Energy}} \bibinfo{volume}{8}, \bibinfo{number}{1} (\bibinfo{year}{2017}), \bibinfo{pages}{425--430}.
\newblock
\href{https://doi.org/10.1109/TSTE.2016.2604082}{doi:\nolinkurl{10.1109/TSTE.2016.2604082}}


\bibitem[Sahoo et~al\mbox{.}(2018)]%
        {sahoo2018}
\bibfield{author}{\bibinfo{person}{Subham~S. Sahoo}, \bibinfo{person}{Christoph~H. Lampert}, {and} \bibinfo{person}{Georg Martius}.} \bibinfo{year}{2018}\natexlab{}.
\newblock \showarticletitle{Learning Equations for Extrapolation and Control}.
\newblock \bibinfo{journal}{\emph{CoRR}}  \bibinfo{volume}{abs/1806.07259} (\bibinfo{year}{2018}).
\newblock
\showeprint[arXiv]{1806.07259}
\urldef\tempurl%
\url{http://arxiv.org/abs/1806.07259}
\showURL{%
\tempurl}


\bibitem[Schmidt and Lipson(2009)]%
        {Schmidt2009}
\bibfield{author}{\bibinfo{person}{M. Schmidt} {and} \bibinfo{person}{H. Lipson}.} \bibinfo{year}{2009}\natexlab{}.
\newblock \showarticletitle{{D}istilling free-form natural laws from experimental data}.
\newblock \bibinfo{journal}{\emph{Science}} \bibinfo{volume}{324}, \bibinfo{number}{5923} (\bibinfo{year}{2009}), \bibinfo{pages}{81--85}.
\newblock


\bibitem[Staelens et~al\mbox{.}(2013)]%
        {Staelens2013}
\bibfield{author}{\bibinfo{person}{Nicolas Staelens}, \bibinfo{person}{Dirk Deschrijver}, \bibinfo{person}{Ekaterina Vladislavleva}, \bibinfo{person}{Brecht Vermeulen}, \bibinfo{person}{Tom Dhaene}, {and} \bibinfo{person}{Piet Demeester}.} \bibinfo{year}{2013}\natexlab{}.
\newblock \showarticletitle{Constructing a No-Reference {H.264/AVC} Bitstream-Based Video Quality Metric Using Genetic Programming-Based Symbolic Regression}.
\newblock \bibinfo{journal}{\emph{IEEE Trans. Cir. and Sys. for Video Technol.}} \bibinfo{volume}{23}, \bibinfo{number}{8} (\bibinfo{date}{Aug.} \bibinfo{year}{2013}), \bibinfo{pages}{1322–1333}.
\newblock
\showISSN{1051-8215}
\href{https://doi.org/10.1109/TCSVT.2013.2243052}{doi:\nolinkurl{10.1109/TCSVT.2013.2243052}}


\bibitem[Topchy et~al\mbox{.}(2001)]%
        {topchy2001}
\bibfield{author}{\bibinfo{person}{Alexander Topchy}, \bibinfo{person}{William~F Punch}, {et~al\mbox{.}}} \bibinfo{year}{2001}\natexlab{}.
\newblock \showarticletitle{Faster genetic programming based on local gradient search of numeric leaf values}. In \bibinfo{booktitle}{\emph{Proceedings of the genetic and evolutionary computation conference (GECCO-2001)}}, Vol.~\bibinfo{volume}{155162}. Morgan Kaufmann San Francisco, CA.
\newblock


\bibitem[Trujillo et~al\mbox{.}(2016)]%
        {Trujillo2016}
\bibfield{author}{\bibinfo{person}{Leonardo Trujillo}, \bibinfo{person}{Luis Munoz}, \bibinfo{person}{Edgar Galvan-Lopez}, {and} \bibinfo{person}{Sara Silva}.} \bibinfo{year}{2016}\natexlab{}.
\newblock \showarticletitle{neat Genetic Programming: Controlling Bloat Naturally}.
\newblock \bibinfo{journal}{\emph{Information Sciences}}  \bibinfo{volume}{333} (\bibinfo{date}{10 March} \bibinfo{year}{2016}), \bibinfo{pages}{21--43}.
\newblock
\showISSN{0020-0255}
\href{https://doi.org/doi:10.1016/j.ins.2015.11.010}{doi:\nolinkurl{doi:10.1016/j.ins.2015.11.010}}


\bibitem[Udrescu and Tegmark(2020)]%
        {Udrescu2020}
\bibfield{author}{\bibinfo{person}{Silviu-Marian Udrescu} {and} \bibinfo{person}{Max Tegmark}.} \bibinfo{year}{2020}\natexlab{}.
\newblock \showarticletitle{{AI Feynman}: A physics-inspired method for symbolic regression}.
\newblock \bibinfo{journal}{\emph{Science Advances}}  \bibinfo{volume}{6} (\bibinfo{date}{04} \bibinfo{year}{2020}), \bibinfo{pages}{eaay2631}.
\newblock
\href{https://doi.org/10.1126/sciadv.aay2631}{doi:\nolinkurl{10.1126/sciadv.aay2631}}


\bibitem[Valipour et~al\mbox{.}(2021)]%
        {valipour21symbolicgpt}
\bibfield{author}{\bibinfo{person}{Mojtaba Valipour}, \bibinfo{person}{Bowen You}, \bibinfo{person}{Maysum Panju}, {and} \bibinfo{person}{Ali Ghodsi}.} \bibinfo{year}{2021}\natexlab{}.
\newblock \showarticletitle{{SymbolicGPT}: {A} Generative Transformer Model for Symbolic Regression}.
\newblock \bibinfo{journal}{\emph{CoRR}}  \bibinfo{volume}{abs/2106.14131} (\bibinfo{year}{2021}).
\newblock
\showeprint[arXiv]{2106.14131}
\urldef\tempurl%
\url{https://arxiv.org/abs/2106.14131}
\showURL{%
\tempurl}


\bibitem[Vastl et~al\mbox{.}(2022)]%
        {Vastl2022}
\bibfield{author}{\bibinfo{person}{Martin Vastl}, \bibinfo{person}{Jonáš Kulhánek}, \bibinfo{person}{Jiří Kubalík}, \bibinfo{person}{Erik Derner}, {and} \bibinfo{person}{Robert Babuška}.} \bibinfo{year}{2022}\natexlab{}.
\newblock \bibinfo{title}{{SymFormer}: End-to-end symbolic regression using transformer-based architecture}.
\newblock
\href{https://doi.org/10.48550/ARXIV.2205.15764}{doi:\nolinkurl{10.48550/ARXIV.2205.15764}}


\bibitem[Verdier et~al\mbox{.}(2019)]%
        {VERDIER2019}
\bibfield{author}{\bibinfo{person}{Cees~F. Verdier}, \bibinfo{person}{Robert Babuška}, \bibinfo{person}{Barys Shyrokau}, {and} \bibinfo{person}{Manuel Mazo}.} \bibinfo{year}{2019}\natexlab{}.
\newblock \showarticletitle{Near Optimal Control With Reachability and Safety Guarantees}.
\newblock \bibinfo{journal}{\emph{IFAC-PapersOnLine}} \bibinfo{volume}{52}, \bibinfo{number}{11} (\bibinfo{year}{2019}), \bibinfo{pages}{230--235}.
\newblock
\showISSN{2405-8963}
\href{https://doi.org/10.1016/j.ifacol.2019.09.146}{doi:\nolinkurl{10.1016/j.ifacol.2019.09.146}}
\newblock
\shownote{5th IFAC Conference on Intelligent Control and Automation Sciences ICONS 2019}.


\bibitem[von Rueden et~al\mbox{.}(2023)]%
        {vonRueden2023}
\bibfield{author}{\bibinfo{person}{Laura von Rueden}, \bibinfo{person}{Sebastian Mayer}, \bibinfo{person}{Katharina Beckh}, \bibinfo{person}{Bogdan Georgiev}, \bibinfo{person}{Sven Giesselbach}, \bibinfo{person}{Raoul Heese}, \bibinfo{person}{Birgit Kirsch}, \bibinfo{person}{Julius Pfrommer}, \bibinfo{person}{Annika Pick}, \bibinfo{person}{Rajkumar Ramamurthy}, \bibinfo{person}{Michal Walczak}, \bibinfo{person}{Jochen Garcke}, \bibinfo{person}{Christian Bauckhage}, {and} \bibinfo{person}{Jannis Schuecker}.} \bibinfo{year}{2023}\natexlab{}.
\newblock \showarticletitle{Informed Machine Learning – A Taxonomy and Survey of Integrating Prior Knowledge into Learning Systems}.
\newblock \bibinfo{journal}{\emph{IEEE Transactions on Knowledge and Data Engineering}} \bibinfo{volume}{35}, \bibinfo{number}{1} (\bibinfo{year}{2023}), \bibinfo{pages}{614--633}.
\newblock
\href{https://doi.org/10.1109/TKDE.2021.3079836}{doi:\nolinkurl{10.1109/TKDE.2021.3079836}}


\bibitem[Werner et~al\mbox{.}(2021)]%
        {werner2021}
\bibfield{author}{\bibinfo{person}{Matthias Werner}, \bibinfo{person}{Andrej Junginger}, \bibinfo{person}{Philipp Hennig}, {and} \bibinfo{person}{Georg Martius}.} \bibinfo{year}{2021}\natexlab{}.
\newblock \showarticletitle{Informed Equation Learning}.
\newblock \bibinfo{journal}{\emph{CoRR}}  \bibinfo{volume}{abs/2105.06331} (\bibinfo{year}{2021}).
\newblock
\showeprint[arXiv]{2105.06331}
\urldef\tempurl%
\url{https://arxiv.org/abs/2105.06331}
\showURL{%
\tempurl}


\bibitem[Willard et~al\mbox{.}(2020)]%
        {Willard2020}
\bibfield{author}{\bibinfo{person}{Jared~D. Willard}, \bibinfo{person}{Xiaowei Jia}, \bibinfo{person}{Shaoming Xu}, \bibinfo{person}{Michael~S. Steinbach}, {and} \bibinfo{person}{Vipin Kumar}.} \bibinfo{year}{2020}\natexlab{}.
\newblock \showarticletitle{Integrating Physics-Based Modeling with Machine Learning: A Survey}.
\newblock \bibinfo{journal}{\emph{ArXiv}}  \bibinfo{volume}{abs/2003.04919} (\bibinfo{year}{2020}).
\newblock


\bibitem[Wilstrup and Kasak(2021)]%
        {Wilstrup2021SymbolicRO}
\bibfield{author}{\bibinfo{person}{Casper Wilstrup} {and} \bibinfo{person}{Jaan Kasak}.} \bibinfo{year}{2021}\natexlab{}.
\newblock \showarticletitle{Symbolic regression outperforms other models for small data sets}.
\newblock \bibinfo{journal}{\emph{ArXiv}}  \bibinfo{volume}{abs/2103.15147} (\bibinfo{year}{2021}).
\newblock


\bibitem[Wu et~al\mbox{.}(2024)]%
        {wu2024prunesymnetsymbolicneuralnetwork}
\bibfield{author}{\bibinfo{person}{Min Wu}, \bibinfo{person}{Weijun Li}, \bibinfo{person}{Lina Yu}, \bibinfo{person}{Wenqiang Li}, \bibinfo{person}{Jingyi Liu}, \bibinfo{person}{Yanjie Li}, {and} \bibinfo{person}{Meilan Hao}.} \bibinfo{year}{2024}\natexlab{}.
\newblock \bibinfo{title}{PruneSymNet: A Symbolic Neural Network and Pruning Algorithm for Symbolic Regression}.
\newblock
\showeprint[arxiv]{2401.15103}~[cs.LG]
\urldef\tempurl%
\url{https://arxiv.org/abs/2401.15103}
\showURL{%
\tempurl}


\bibitem[Zhou and Pan(2022)]%
        {Zhou2022}
\bibfield{author}{\bibinfo{person}{Hongpeng Zhou} {and} \bibinfo{person}{Wei Pan}.} \bibinfo{year}{2022}\natexlab{}.
\newblock \bibinfo{title}{Bayesian Learning to Discover Mathematical Operations in Governing Equations of Dynamic Systems}.
\newblock
\href{https://doi.org/10.48550/ARXIV.2206.00669}{doi:\nolinkurl{10.48550/ARXIV.2206.00669}}


\bibitem[Žegklitz and Pošík(2019)]%
        {zegklitz2019}
\bibfield{author}{\bibinfo{person}{Jan Žegklitz} {and} \bibinfo{person}{Petr Pošík}.} \bibinfo{year}{2019}\natexlab{}.
\newblock \showarticletitle{Symbolic regression in dynamic scenarios with gradually changing targets}.
\newblock \bibinfo{journal}{\emph{Applied Soft Computing}}  \bibinfo{volume}{83} (\bibinfo{year}{2019}), \bibinfo{pages}{105621}.
\newblock
\showISSN{1568-4946}
\href{https://doi.org/10.1016/j.asoc.2019.105621}{doi:\nolinkurl{10.1016/j.asoc.2019.105621}}


\end{thebibliography}
